\documentclass{article}

\usepackage{iclr2027_conference}
\usepackage{iftex}
\ifPDFTeX
  \usepackage{times}
\else
  \usepackage[T1]{fontenc}
  \usepackage{newtxtext}
\fi

\usepackage{amsmath,amsfonts,bm}

\def\eqref#1{equation~\ref{#1}}

\def\1{\bm{1}}

\DeclareMathAlphabet{\mathsfit}{\encodingdefault}{\sfdefault}{m}{sl}
\SetMathAlphabet{\mathsfit}{bold}{\encodingdefault}{\sfdefault}{bx}{n}

\usepackage{float}
\definecolor{citecolor}{HTML}{0071bc}
\usepackage[pagebackref=false,breaklinks=true,colorlinks,citecolor=citecolor,bookmarks=false]{hyperref}
\usepackage{url}

\usepackage{amsthm}
\usepackage{amssymb}
\usepackage{booktabs}
\usepackage{graphicx}
\usepackage{colortbl}
\definecolor{ourrow}{HTML}{EBF5FB}
\definecolor{cfrcgreen}{HTML}{196F3D}
\definecolor{cfrcred}{HTML}{922B21}
\definecolor{cfrcblue}{HTML}{1B4F72}
\usepackage{tabularx}
\usepackage{enumitem}
\usepackage{cleveref}
\crefname{equation}{Eq.}{Eqs.}
\makeatletter
\renewcommand{\eqref}[1]{\textup{(\ref{#1})}}

\title{Do Not Restart: Residual Completion for Stateful Agent Handoffs}
\author{
Runzhi Deng$^{1,2}$\thanks{Equal contribution.},\hspace{0.5em}
Yiming Zhong$^{1}$\footnotemark[1],\hspace{0.5em}
Fang Zhao$^{1}$\thanks{Corresponding authors.},\hspace{0.5em}
Pan Zhou$^{2}$\footnotemark[2] \\
$^{1}$Nanjing University \qquad $^{2}$Singapore Management University
}

\newcommand{\cfrc}{\textsc{CFRC}}

\newtheorem{proposition}{Proposition}

\iclrfinalcopy

\begin{document}
\raggedbottom

\maketitle
\lhead{}
\begin{abstract}

Routing and cascades reduce tool-agent cost by transferring control across models, but stateful handoffs must preserve accepted choices, realized effects, and unfinished obligations. We formulate this as \emph{commitment-constrained residual completion} and introduce Commitment-Frontier Residual Completion (\cfrc{}). \cfrc{} enforces target-before-proposal, whole-proposal-before-authority, and live-evidence-before-success: it freezes a residual contract from accepted progress, closes the successor continuation into an evidence-linked graph, and admits execution only when the remainder is covered, with live receipts discharging obligations. We establish contract-relative partial correctness, which extends to the original residual request under complete contract construction. Across five environments and two same-provider model pairs, \cfrc{} achieves comparable macro accuracy to strong full-task agents at 22.0\%--34.6\% mean per-surface cost, with additional cross-provider results demonstrating broader transfer.

\end{abstract}
\section{Introduction}
\label{sec:introduction}
Large language models (LLMs) and multimodal agents increasingly serve as the decision-making core of tool-using systems for planning, software interaction, and multi-step execution \citep{trivedi2024appworld,lu2024toolsandbox}.
Such trajectories repeatedly invoke a model for reasoning, action selection, tool use, and verification, making it costly and often unnecessary to use the strongest model throughout.
A practical alternative is to let lower-cost models handle routine steps and transfer control when tasks become difficult or uncertain \citep{ong2024routellm,son2026swerouter}.
Its benefit, however, depends on whether the successor can reuse accepted progress rather than restart.
Unlike stateless prompt routing \citep{chen2023frugalgpt}, agents operate in persistent environments where actions can produce irreversible side effects.
Restarting wastes computation and risks duplicate actions, while continuing without clear specifications risks goal drift.
This raises a broader handoff question: when control passes between models, how can the successor continue from the reached state while preserving accepted choices and realized effects, completing open obligations, and avoiding duplicate actions?

\begin{figure}[t]
  \centering
  \vspace{-1.5em}
  \includegraphics[width=\textwidth]{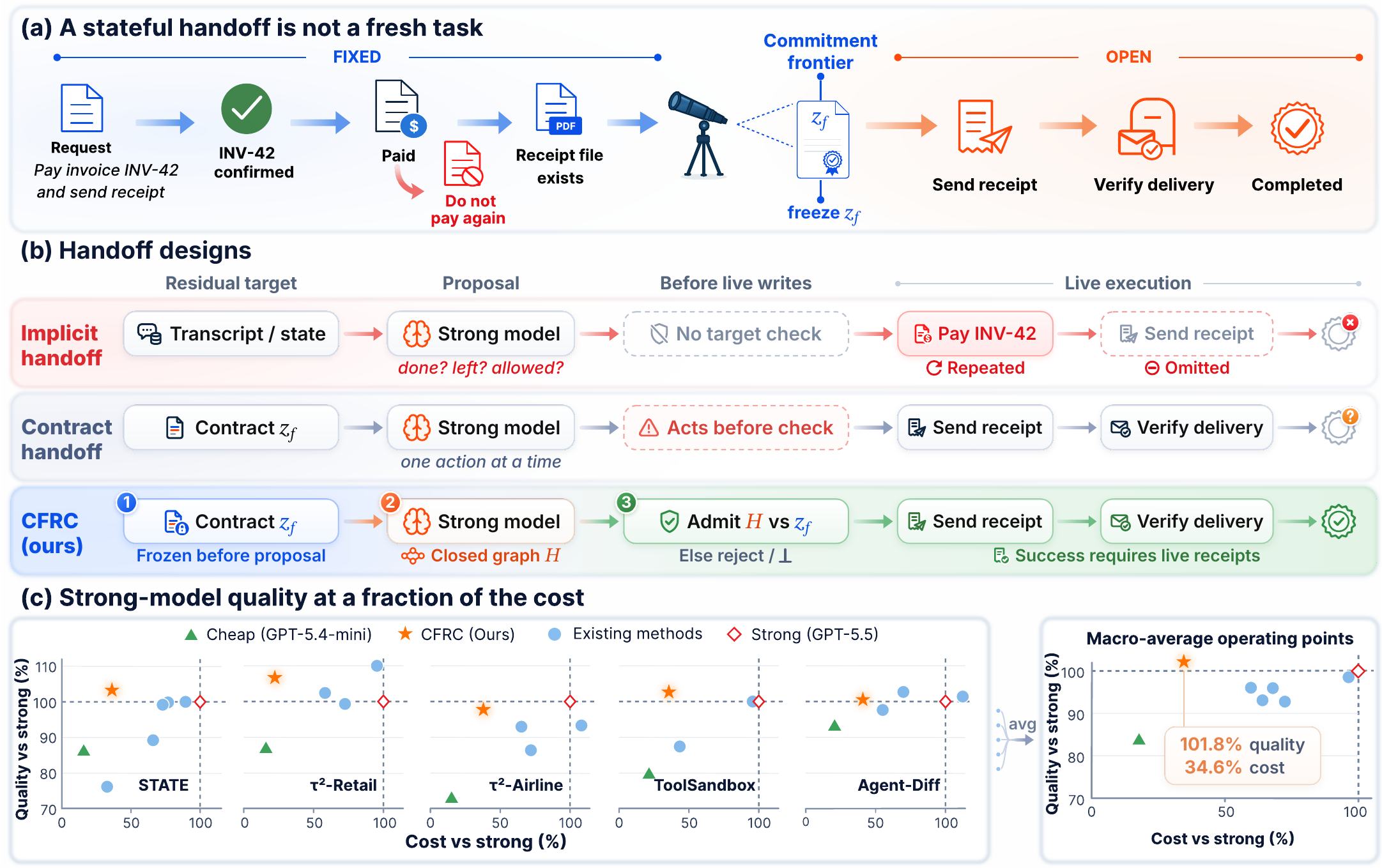}
  \vspace{-1.5em}
  \caption{
  	\textbf{Stateful handoff is residual completion, not restart.}
  	(a) At a commitment frontier, accepted progress is fixed while
  	request-scoped work remains open: INV-42 has been confirmed and paid,
  	so the successor must preserve that effect, deliver its receipt, and
  	verify delivery rather than repay or restart.
  	(b) Implicit handoff can miss open work or repeat completed effects;
  	a contract-only handoff supplies a residual target but still lets the
  	successor act before the whole remainder is checked.
  	\cfrc{} instead freezes the target before proposal, closes the
  	successor's full remainder into an evidence-linked graph before native
  	authority, and accepts success only from live evidence.
  	(c) For GPT-5.4-mini/GPT-5.5, quality and cost are normalized to the
  	strong full-task agent on each surface $(100\%,100\%)$.
  	Across five surfaces, \cfrc{} remains near strong-model quality and
  	reaches 101.8\% of strong macro accuracy at a 34.6\% mean cost ratio.
  }
  \label{fig:motivation}
  \vspace{-0.2em}
\end{figure}

In the invoice task of Fig.~\ref{fig:motivation}a, the user asks the agent to pay one outstanding invoice and send its receipt.
The first model selects an illustrative invoice \texttt{INV-42}, which the user confirms, making the choice binding even before any tool call.
At a later handoff, payment has succeeded but receipt delivery remains pending.
 The successor cannot restart, as another API payment call would cause a duplicate charge, nor switch invoices, since the user confirmed \texttt{INV-42}.
	Yet declaring success is also wrong because delivery remains unfulfilled.
	The task is therefore neither fresh nor complete: the accepted trajectory contains fixed progress alongside unfinished work. 
We call this handoff point a \emph{commitment frontier} and frame the challenge as \emph{commitment-constrained residual completion}.
Under this formulation,  a valid continuation must execute from the reached state, preserve accepted choices and realized effects, and complete all open request-scoped obligations.

Existing agent components address adjacent parts of this problem but leave three gaps, as illustrated in Fig.~\ref{fig:motivation}b.
First, routing, handoff, and memory mechanisms determine who acts or transfer past context \citep{kc2026handoff,wang2024awm}, but they do not independently specify what the successor must still preserve and complete.
In the invoice example, transferring the transcript or reached state alone does not prevent the successor from overlooking delivery or reinterpreting the confirmed choice.
Second, planners and transcript-conditioned successors can propose later actions \citep{yao2023react}, but a proposal cannot reveal what it omits.
A continuation containing only a status check and success response provides no independent signal that receipt delivery remains missing.
Third, action-level guards and control layers can reject unsafe or invalid proposed calls \citep{wang2025agentspec,kamath2025agentc,liu2026toolgate}, such as duplicate payments or unauthorized invoices.
Yet they cannot require an action that was never proposed, nor establish whole-suffix completeness before incremental native execution begins.
Thus, existing mechanisms address context transfer, action proposal, or local validity, but not the joint need for an independent residual target, complete proposal coverage, and evidence-backed realization.

\paragraph{Our approach.}
We address this gap with Commitment-Frontier Residual Completion (\cfrc{}).
The method enforces target before proposal, whole proposal before authority, and live evidence before success, as illustrated in Fig.~\ref{fig:motivation}b.
First, Reached-State Contract Construction in \cfrc{} derives a frozen residual contract $z_f$ from the accepted trajectory, reached state, trusted receipts, and public tool interfaces before seeing the successor’s proposal.
The contract records what is fixed and what remains open, so omitted work cannot disappear merely because the successor fails to mention it.
Second, Executable Residual Completion in \cfrc{} lets the successor author on an isolated replica where available and closes the full continuation as an evidence-linked graph $H$, exposing its actions, dependencies, and obligation coverage without changing the live state.
Finally, Checked Execution in \cfrc{} admits the complete $H$ against $z_f$ before granting native authority, rebinds dynamic values to live predecessor receipts, and accepts completion only when matching native evidence discharges every obligation.
In the invoice example, $z_f$ preserves the completed payment while recording pending delivery; the  graph $H$ must explicitly include that delivery path.
Checked Execution rejects duplicate or incomplete proposals before live execution and confirms success only after live receipts verify delivery. Finally, we theoretically 
establish conditional partial correctness, showing that successfully realized prefixes preserve encoded progress and returned suffixes satisfy the contract. 

Across five stateful tool-use surfaces, \cfrc{} shifts the quality--cost
frontier (Fig.~\ref{fig:motivation}c). With GPT-5.4-mini/GPT-5.5, it achieves
70.8\% macro accuracy versus 69.6\% for GPT-5.5 at 34.6\% mean cost; with
Luna/Sol, it achieves 70.8\% versus 70.0\% at 22.0\% mean cost. Cross-provider
Luna/Sonnet~5 and Luna/Gemini~3.7~Flash remain within 0.5 points of their
strong anchors at 26.0\% and 44.1\% cost. Checked Execution blocks invalid
proposals before live writes, while an independent audit recovers 62/68
persistent write obligations and finds no ungrounded writes among 50 admitted
proposals. So \cfrc{} preserves strong-model quality while focusing
expensive reasoning on unfinished work.

 \vspace{-0.7em}
\section{Related Work}
\label{sec:related-work}
 \vspace{-0.5em}
 \textbf{Model routing \& agent handoff.}
Planning and reflection improve how agents generate actions \citep{yao2023react,shinn2023reflexion,wang2024awm}.
Routers and cascades select models by predicted difficulty or cost \citep{ong2024routellm,bouchard2026escalation,son2026swerouter,wei2026steplevel,liu2026agenticrouting}, while handoff frameworks quantify information loss across checkpoints \citep{kc2026handoff}.
Yet a transferred transcript does not specify which effects are fixed or which obligations remain open. \cfrc{} constructs this residual target before the successor proposes a continuation.

\textbf{Classical specifications \& recovery.}
Conditional correctness relates execution to explicit pre- and postconditions~\citep{hoare1969axiomatic}.
Sagas structure long-lived transactions around completion or compensation~\citep{garciamolina1987sagas}, while plan repair adapts existing plans to changed execution contexts~\citep{fox2006planstability}.
\cfrc{} follows this separation of specification and execution, but constructs the residual target from accepted model-generated progress to constrain the successor's entire proposal. Accepted effects are preserved, while a failed suffix may return $\bot$ without compensation.

\textbf{Stateful control \& verification.}
Runtime guards, capabilities, and transactional frameworks constrain actions
outside the model
\citep{wang2025agentspec,kamath2025agentc,debenedetti2025camel,fan2026pact,
	liu2026toolgate,lin2026vigil,patil2024goex,chen2026cordon}.
Execution-monitoring limits are classical~\citep{schneider2000enforceable}.
Our omission argument instead concerns verifying planned coverage
\emph{before live writes}: an incomplete prefix cannot reveal which obligations
a future continuation will omit, although explicit terminal obligations can
still detect false completion. \cfrc{} checks the entire proposed remainder
against a separately constructed target and requires live evidence for success. Stateful benchmarks
\citep{yao2024taubench,barres2025tau2bench,lu2024toolsandbox,trivedi2024appworld,
	microsoft2026statebench,pysklo2026agentdiff} make the gap between attempted
actions and supported completion measurable
\citep{advani2026falsesuccess,gao2026evidencebounds}.

\vspace{-0.5em}
\section{Methodology}
\label{sec:method}
\vspace{-0.5em}

\subsection{Stateful Handoff as a Residual-Completion Problem}
\label{sec:method-setting}
\vspace{-0.5em}

\textbf{Problem setup.}
A stateful handoff occurs when one model takes over an ongoing tool-use
	trajectory produced by another, as in routed or cascaded agents for coding,
	web navigation, and other applications with persistent state or externally
	visible effects. Unlike a fresh request, the successor inherits both
	\emph{fixed progress} that must be preserved and \emph{open work} that remains
	to be completed. We study how to finish this residual work without replacing,
	undoing, or duplicating accepted progress.

Formally, let $q$ be the original request, $\tau_t$ the accepted trajectory
	prefix, $s_t$ the reached live state, and $\mathcal R_t$ the trusted native
	receipts available at $s_t$. Let $\mathcal A_{\mathrm{env}}$ denote the public
	tool interfaces, and let $\mathcal O_t$ denote the semantic, request-scoped obligation
	instances still open under $q$ and $\tau_t$. We use $B_t$ for accepted choice
	bindings, including user-confirmed choices that may precede any native effect,
	and $K_t$ for realized effects supported by $\mathcal R_t$. With fixed progress
	$F_t=(B_t,K_t)$, residual completion seeks a continuation $\sigma$ from $s_t$
	satisfying
\begin{equation}
	\mathrm{Exec}(\sigma;s_t)
	\;\wedge\;
	\mathrm{Preserve}(\sigma;F_t)
	\;\wedge\;
	\mathrm{Complete}(\sigma;\mathcal O_t).
	\label{eq:residual-objective}
\end{equation}
Here, the three predicates require the suffix to execute from the reached
	state, respect accepted bindings and realized effects, and discharge every
	open obligation before claiming success. If \cfrc{} cannot establish such a
	suffix within budget, it returns $\bot$; this does not imply that no valid
	continuation exists. For the formal guarantee, each in-scope obligation has a request-defined,
	finitely checkable predicate over the resulting trace and state, independent
	of the contract later constructed by \cfrc{}. Let $\Sigma_f(q,\tau_t)$ denote
	the continuations satisfying Eq.~\eqref{eq:residual-objective}.

\begin{figure}[t]
	\centering
	\includegraphics[width=\textwidth]{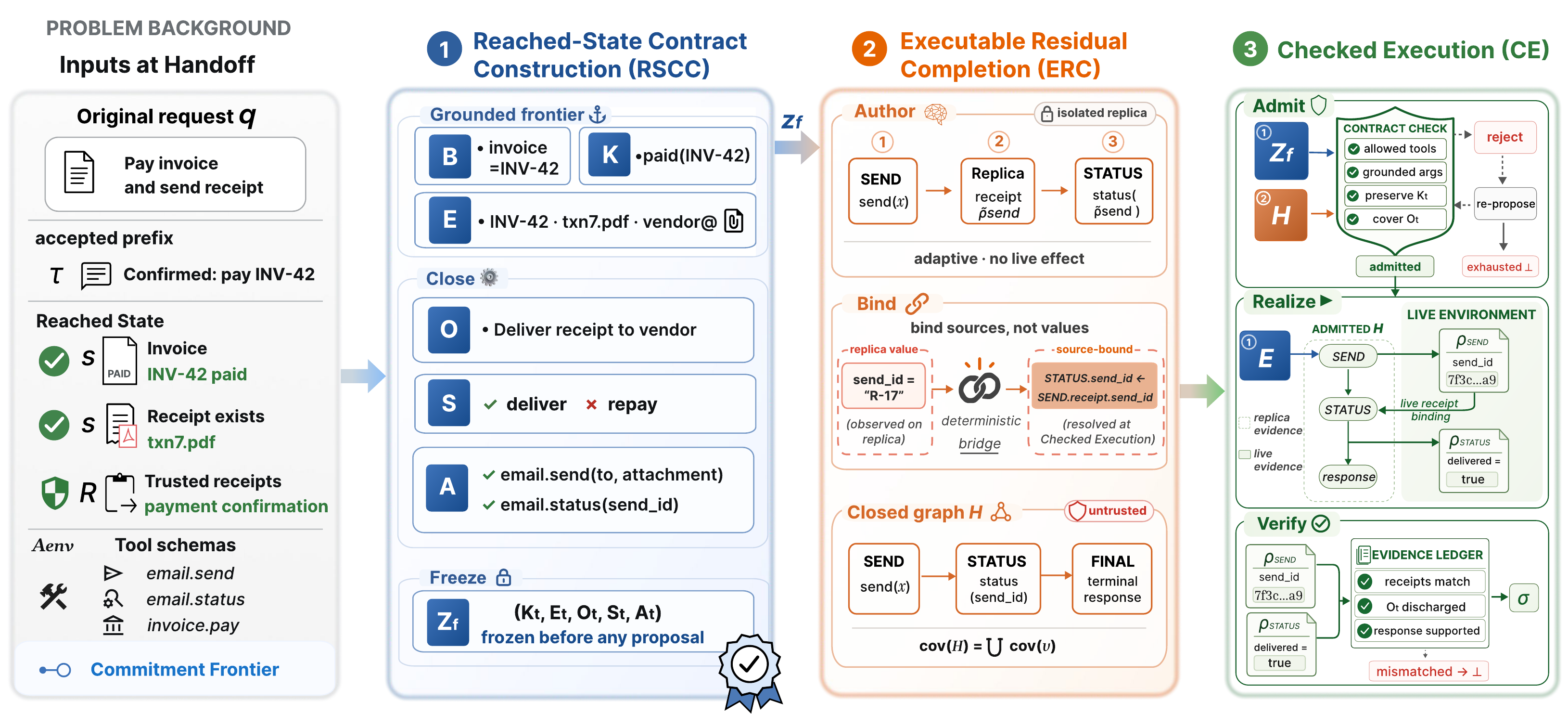}
	\caption{\textbf{\cfrc{} at a post-payment commitment frontier.}
			RSCC identifies a supported frontier and freezes the residual contract
			$z_f$ before querying the successor. ERC then closes the proposed remainder
			as an evidence-linked graph $H$, and CE admits and realizes $H$, returning
			$\sigma$ only when matching live evidence discharges every encoded open
			obligation, and $\bot$ otherwise.}
	\label{fig:overview}
\end{figure}

\textbf{Running example and commitment frontier.}
Consider the post-payment invoice handoff in Fig.~\ref{fig:overview}. The
	request $q$ asks the agent to pay one outstanding invoice and send its receipt;
	$\tau_t$ records that \texttt{INV-42} was selected and confirmed and that its
	payment succeeded; $s_t$ contains the paid invoice and receipt file
	\texttt{txn7.pdf}; and $\mathcal R_t$ contains the trusted native evidence.
	Thus, $B_t$ fixes $\texttt{invoice}=\texttt{INV-42}$, $K_t$ contains the
	receipt-backed effect $\mathrm{paid}(\texttt{INV-42})$, and $\mathcal O_t$
	still requires delivery of the receipt to the verified recipient. A valid
	$\sigma$ may therefore \emph{send receipt $\rightarrow$ verify delivery
		$\rightarrow$ confirm success}: repaying violates preservation, omitting
	delivery violates completion, and unsupported success violates
	Eq.~\eqref{eq:residual-objective}. The same issue can arise before any change to the environment: once the user confirms
	\texttt{INV-42}, that binding already constrains future behavior. We call such
	a point a \emph{commitment frontier}: the accepted prefix has fixed progress that must be
	preserved while request-scoped work remains open—in short, \emph{fixed
		progress plus unfinished work}. Before any accepted choice or realized effect,
	the task is fresh; after all obligations are complete, no residual task remains.

\textbf{From challenges to \cfrc{}.}
Eq.~\eqref{eq:residual-objective} exposes three design challenges.
	\textbf{(C1) What is fixed, and what remains?} Neither the request nor the
	reached state alone necessarily specifies the residual task, while letting the
	successor define its own target makes omissions difficult to detect. \cfrc{}
	therefore constructs an independent frozen contract $z_f$ before proposal
	(Sec.~\ref{sec:contract-construction}).
	\textbf{(C2) Does the proposed remainder cover that target?} Call-local checks
	can reject an invalid action but cannot detect an obligation for which no action
	is proposed. In stateful environments with irreversible tool mutations,
	step-by-step execution risks leaving the environment in an unrecoverable state
	before discovering that downstream obligations lack valid inputs. \cfrc{} therefore
	exposes the whole remainder as a closed graph $H$ before native execution
	(Sec.~\ref{sec:residual-completion}).
	\textbf{(C3) Were the proposed actions actually realized?} Replica results are only
	proposal-time evidence, and native calls may fail. \cfrc{} therefore grants
	native authority only after validation and accepts completion only from
	matching live evidence (Sec.~\ref{sec:checked-execution}). These three stages form the pipeline in Fig.~\ref{fig:overview}:
\begin{equation}
	(q,\tau_t,s_t,\mathcal R_t,\mathcal A_{\mathrm{env}})
	\;\longrightarrow\;
	z_f
	\;\longrightarrow\;
	H
	\;\longrightarrow\;
	\sigma\ \text{or}\ \bot.
	\label{eq:cfrc-flow}
\end{equation}
We study one handoff tenure at a time. \cfrc{} triggers handoff through a fixed
	supported-frontier rule, but does not optimize general model routing or choose
	among multiple successors; it also preserves supported prior commitments
	rather than re-evaluating their desirability.

\subsection{Reached-State Contract Construction}
\enlargethispage{-\baselineskip}
\label{sec:contract-construction}
\vspace{-0.5em}

Reached-State Contract Construction (RSCC) addresses C1 in Sec.~\ref{sec:method-setting} by constructing the
	residual target \emph{before} the successor proposes anything. Rather than
	merely summarize the trajectory, RSCC converts accepted history into a
	proposal-independent specification of what is fixed and what remains owed. It
	\textbf{grounds} accepted progress, \textbf{closes} the remaining request, and
	\textbf{freezes} the resulting contract $z_f$. RSCC uses only
	$q,\tau_t,s_t,\mathcal R_t$ and a fixed adapter over public tool schemas. The
	adapter maps public interfaces into shared argument, effect, and evidence
	types; it cannot access simulator internals, reference outcomes, or evaluator
	rewards.

\textbf{Step 1: Grounding fixed progress.}
RSCC first identifies the binding parts of the accepted history:
\[
(B_t,K_t,E_t)
=
\operatorname{Ground}(\tau_t,s_t,\mathcal R_t),
\]
where $B_t$ contains accepted choice bindings, $K_t$ contains receipt-backed
realized effects that later actions must preserve, and $E_t$ contains verified
entities, values, files, and receipt fields that the residual continuation may
safely reference. Grounding is deterministic: public tool schemas specify mutating
effects and receipt formats, while $\operatorname{Ground}$ parses trusted native
receipts in $\mathcal R_t$ to populate $K_t$ and extracts confirmed entities into
$E_t$ and $B_t$. In our invoice example, $E_t$ retains \texttt{txn7.pdf} and the
verified recipient, while $K_t$ records the completed payment.

\textbf{Step 2: Closing the residual task.}
Grounding identifies what is fixed but not what remains. RSCC compares the
grounded progress with the original request $q$ and the public interfaces to derive
\begin{equation}	\label{eq:contract}
	\begin{split}
		(\tau_t,s_t,\mathcal R_t)
		&\xrightarrow{\operatorname{Ground}}
		(B_t,K_t,E_t)
		\xrightarrow[\mathcal A_{\mathrm{env}}]{q,\operatorname{Close}}
		z_f=(K_t,E_t,O_t,\mathcal S_t,\mathcal A_t).
	\end{split}
\end{equation}
Here, discrete $O_t = \operatorname{canon}_q(\mathcal O_t)$ encodes semantic obligations
$\mathcal O_t$; $\mathcal S_t$ specifies admissible residual effects, and
$\mathcal A_t$ the interfaces that may realize or verify them.
Closure operationalizes the division of labor between rules and model inference.
On structured surfaces (STATE-Bench, ToolSandbox, Agent-Diff), candidate obligations
are extracted deterministically by matching request intents against tool schemas
and excluding effects already certified in $K_t$.
On conversational surfaces ($\tau^2$), an extraction prompt decomposes user request
$q$ relative to prefix $\tau_t$ into candidate obligations $O_t$, parameterizing
them with entities from $E_t$ and restricting them to admissible interfaces
$\mathcal A_t$.
For the post-payment invoice, $O_t$ encodes delivery of \texttt{txn7.pdf} to the
verified recipient, $\mathcal S_t$ permits delivery and receipt-backed confirmation
while excluding repayment, and $\mathcal A_t$ retains the corresponding send and status
interfaces.
Accepted bindings in $B_t$ are compiled into fixed values in $E_t$, obligations
in $O_t$, and restrictions in $\mathcal S_t$, while receipt-backed effects in
$K_t$ remain protected against reversal or repetition.

\textbf{Step 3: Validating and freezing the contract.}
	RSCC requires every contract field to have trusted support, $O_t$ to be
	nonempty, and the accepted prefix to contain at least one supported accepted binding or
	receipt-backed realized effect. Once validated, $z_f$ is frozen
	\emph{before} querying the successor: neither proposal nor execution may expand, weaken, or rewrite it, so
	recorded work cannot disappear because the successor omits it. These checks
	establish support but do not prove that RSCC captured every semantic obligation
	in $\mathcal O_t$. Freezing prevents \emph{proposal-induced} omission; it cannot
	recover an obligation missed during construction. Accordingly, the lift from
	contract-relative correctness to the original residual request is conditional
	on complete RSCC construction (Sec.~\ref{sec:guarantee}). If construction or
	validation fails, \cfrc{} withholds successor write authority and returns
	$\bot$.

\vspace{-0.4em}
\subsection{Executable Residual Completion}
\label{sec:residual-completion}
\vspace{-0.4em}
The frozen contract $z_f$ specifies \emph{what} must be preserved and completed while
leaving \emph{how} to the successor. ERC addresses C2 in Sec.~\ref{sec:method-setting} by exposing the
successor's complete proposed remainder before any new live write:
the successor \textbf{authors} a continuation, a deterministic bridge
\textbf{binds} its data and evidence dependencies, and the proposal is
\textbf{closed} as an untrusted graph $H$.

\textbf{Step 1: Data-dependent authoring without live effects.}
	Residual plans are often data-dependent because later call arguments depend on
	earlier outputs. When safe snapshotting is available, the successor authors against an
	isolated replica of $s_t$, observing replica receipts $\widetilde{\rho}_v$
	without changing the live environment. On an unforkable surface, the successor
	instead closes the graph from handoff-time evidence and symbolic receipt
	dependencies; no new state-changing live call is authorized before CE in Sec.~\ref{sec:checked-execution} admits
	the complete proposal. Replica receipts may expose intermediate values for
	authoring, but they neither prove live effects nor discharge obligations. For
	example, the successor may send \texttt{txn7.pdf} on a replica, observe a SEND
	receipt, and use its identifier to propose a later delivery-status query
	without sending a real email.

\textbf{Step 2: Binding dependencies and closing the residual graph.}
	A proposal authored from replica values cannot simply be replayed because
	replica-generated identifiers may differ from live ones.

\pagebreak
\noindent A deterministic bridge
	therefore builds an evidence-linked candidate graph. Let $H$ denote the closed
	DAG retained after dependency-closed candidate selection, containing typed
	calls, the terminal response, arguments, dependencies, source bindings,
	obligation coverage, effects, and required authority.

	Crucially, dynamic arguments bind to their \emph{source}, not to a proposal-time value.
	For example, the status query refers to the \texttt{send\_id} field of the
	preceding SEND receipt; CE (Sec.~\ref{sec:checked-execution}) later resolves that field from the corresponding
	live receipt. The successor chooses the actions, task-specific arguments,
	semantic order, and terminal response. The bridge only records or normalizes
	structure supported by $z_f$, public schemas, and available receipts; it does
	not choose another task solution.

\textbf{Step 3: Closing the whole remainder without repair.}
Unlike next-action control, $H$ exposes the \emph{entire} proposed remainder.
Writing $\operatorname{cov}(v)$ for the obligations claimed by node $v$,
{\setlength{\abovedisplayskip}{3pt plus 1pt minus 1pt}\setlength{\belowdisplayskip}{3pt plus 1pt minus 1pt}
\begin{equation}
	\operatorname{cov}(H)
	=
	\bigcup\nolimits_{v\in V(H)}\operatorname{cov}(v),
	\label{eq:graph-coverage}
\end{equation}}
can later be compared with $O_t$. Thus an omitted action becomes visible as
missing coverage: an invoice proposal without a SEND node leaves receipt
delivery uncovered.

	Crucially, while a terminal monitor can detect omission at episode end, it cannot roll back irreversible tool mutations already committed to the live environment.
	Nor can step-by-step checkers observe whether future actions will complete open obligations.
	Closing $H$ beforehand validates full obligation coverage and dataflow reachability \emph{before} granting live write authority, preventing destructive partial execution.
The bridge does not repair the proposal; it normalizes supported
dependencies, source bindings, or coverage annotations without adding,
removing, or replacing task actions. The resulting $H$ is inspectable but untrusted.

\vspace{-0.4em}
\subsection{Checked Execution}

\label{sec:checked-execution}
\vspace{-0.4em}

The closed $H$ from Sec.~\ref{sec:residual-completion} makes the successor's proposal explicit but does not establish
that it is contract-valid or that its claimed effects actually occur.
Checked Execution (CE) resolves C3 by putting
\emph{state-changing live authority}
behind whole-graph admission and accepting completion only from live evidence.
CE first \textbf{admits} $H$ against $z_f$, then \textbf{realizes} admitted nodes
using live receipt bindings, and finally \textbf{verifies} that the resulting
evidence discharges every obligation.

\textbf{Step 1: Whole-graph admission before live write authority.}
Before the first state-changing native call, CE compares the fixed proposal
$H$ with
$z_f=(K_t,E_t,O_t,\mathcal S_t,\mathcal A_t)$.
Admission requires:
(i) calls use allowed interfaces with well-typed, grounded arguments;
(ii) proposed effects stay within $\mathcal S_t$ and preserve $K_t$ and the
accepted bindings encoded by the contract;
(iii) the entire graph covers the encoded obligations,
$\operatorname{cov}(H)=O_t$; and
(iv) dependencies and terminal claims have explicit evidence paths.
These conditions correspond to proposal-level executability, preservation, and
coverage. Because CE sees the complete $H$, it can reject duplicate actions,
uncovered obligations, unresolved dependencies, or unsupported terminal claims
before the first live write.
Whole-graph admission establishes that the \emph{planned} suffix satisfies the
contract structure; it does not imply that native execution will succeed.
CE admits or rejects the selected $H$ as a whole. Compiler selection may prune
proposal nodes only while retaining dependencies and full coverage of $O_t$;
it cannot synthesize repair actions.

\textbf{Step 2: Receipt-bound live realization.}
	An admitted graph executes in dependency order on the live environment. Static
	inputs resolve from $E_t$, whereas dynamic inputs are rebound to the
	corresponding \emph{live} predecessor receipts specified by $H$. After each
	call, CE checks the returned receipt against the result and effect instance
	recorded for that node before its fields may feed downstream calls. Replica and
	proposal-time outputs never substitute for native evidence.

\textbf{Step 3: Evidence-backed completion.}
Replica success never discharges an obligation.
Let $s_j$ and $\mathcal R_j$ denote the live state and matching native receipts
after $j$ successfully realized action nodes. CE tracks the obligations whose
evidence remains valid at that point:
{\setlength{\abovedisplayskip}{2pt plus 1pt minus 1pt}\setlength{\belowdisplayskip}{2pt plus 1pt minus 1pt}
\begin{equation}
	L_j
	=
	\{\,o\in O_t:
	\operatorname{wit}_j(o,\mathcal R_j,s_j)\,\},
	\qquad L_0=\varnothing,
	\label{eq:obligation-ledger}
\end{equation}}
where $\operatorname{wit}_j$ requires instance-matching native evidence whose
discharge remains valid at $s_j$. Event obligations may remain discharged once
witnessed; mutable-state obligations require current-state validity. Recorded
invalidation removes discharge and blocks completion for the current tenure;
a fresh receipt cannot discharge that obligation again under the same contract.

If $m$ action nodes are realized, CE returns a suffix $\sigma$ only when every
	node has a matching receipt, $L_m=O_t$, and the terminal response is supported
	by the corresponding evidence:
{\setlength{\abovedisplayskip}{3pt plus 1pt minus 1pt}\setlength{\belowdisplayskip}{3pt plus 1pt minus 1pt}
\begin{equation}
	\Gamma_{z_f}(H)
	=
	\begin{cases}
		\sigma, & \text{if admission and all live-evidence checks succeed}, \\
		\bot, & \text{otherwise}.
	\end{cases}
	\label{eq:ce-decision}
\end{equation}}
A rejection before live execution may trigger another complete proposal
against the same frozen contract within budget.
Once a state-changing live call has occurred, however, a failure or mismatched
receipt ends the current tenure; CE does not repair the old graph or continue
under stale contract evidence. A later attempt reconstructs a contract from the
new reached state. Importantly, $\bot$ after a live call does \emph{not} imply
rollback or that the environment remained unchanged.

\vspace{-0.4em}
\subsection{Conditional End-to-End Guarantee}
\label{sec:guarantee}
\vspace{-0.4em}

The three stages establish complementary properties:
Under complete construction, RSCC faithfully specifies the residual task;
ERC exposes the successor's proposal without repairing it, and CE faithfully
realizes an admitted proposal using live evidence.

\begin{proposition}[Contract-relative partial correctness and residual-task lift]
	\label{prop:end-to-end}
   Let $P$ be a successor-authored residual proposal and
		$H=\operatorname{ERC}_{z_f}(P)$ its closed graph. Under faithful ERC
		closure, sound schemas and checks, valid live evidence, and tenure isolation,
		if
		$\Gamma_{z_f}(H)=\sigma\neq\bot$, then every successfully realized prefix
		preserves the frozen progress and
	{\setlength{\abovedisplayskip}{2pt plus 1pt minus 1pt}\setlength{\belowdisplayskip}{2pt plus 1pt minus 1pt}
	\begin{equation}
		\operatorname{cov}(H)=O_t,
		\qquad
		\sigma\in\Sigma(z_f).
		\label{eq:end-to-end-guarantee}
	\end{equation}}
  If RSCC satisfies the complete-construction assumption, then
		$\sigma\in\Sigma_f(q,\tau_t)$ and satisfies
		Eq.~\eqref{eq:residual-objective}.
\end{proposition}

Proposition~\ref{prop:end-to-end} separates execution fidelity from
	specification fidelity. Under the assumptions stated in Proposition~\ref{prop:end-to-end}, ERC and CE ensure that the returned suffix
	satisfies the frozen contract: the closed graph covers every encoded
	obligation, each successfully realized prefix preserves encoded progress, and
	live evidence supports completion. Under complete RSCC construction, this contract-relative result lifts to the original in-scope residual request. The returned suffix therefore executes from the reached state,
	preserves accepted progress, and
	completes every in-scope semantic obligation still open at handoff. The result
	does not certify the constructor, guarantee that a valid proposal will be
	found, or imply rollback when $\bot$ is returned.

\vspace{-0.4em}
\section{Experiments}
\label{sec:experiments}
\vspace{-0.4em}

Our evaluation investigates four questions: whether \cfrc{} approaches strong-model quality at lower inference cost, whether these gains hold across model pairs and providers, what RSCC, ERC, and CE individually contribute, and whether independent audits validate the contract and execution mechanisms.

\textbf{Surfaces and protocol.}
We evaluate across five stateful tool-use benchmarks, comprising STATE-Bench,
$\tau^2$-Retail, $\tau^2$-Airline, ToolSandbox, and Agent-Diff
\citep{microsoft2026statebench,barres2025tau2bench,lu2024toolsandbox,
	pysklo2026agentdiff}. All methods use the same frozen tasks, public tool
environment, and scorer. \cfrc{} accesses only public action schemas, effect
types, and native receipts, without simulator internals, reference outputs, or
evaluator rewards.

\textbf{Models and baselines.}
Our primary evaluations use GPT-5.4-mini/GPT-5.5 and GPT-5.6 Luna/Sol; three
	cross-provider pairs test transfer. Baselines include cheap and strong
	full-task agents, StepWise, MTRouter, RouteLLM-BERT, FrugalGPT, and
	AgServe-QAC
	\citep{wei2026steplevel,zhang2026mtrouter,ong2024routellm,
		chen2023frugalgpt,ren2025agserve}. Policies and operating points are fixed
	before test evaluation.

\textbf{Metrics and cost.}
We report official task success except for ToolSandbox, whose metric is
strict all-milestone completion. Cost is cache-neutral provider expenditure over all agent calls,
excluding user-simulator and evaluator roles. Aggregates are computed over each
method's evaluated surfaces.

\subsection{End-to-End Quality and Cost}
\label{sec:main-results}

\begin{table}[t]
	\centering
	\scriptsize
	\setlength{\tabcolsep}{2.5pt}
	\renewcommand{\arraystretch}{0.80}
	\caption{End-to-end results. Scores are task success (\%), except that
			ToolSandbox uses strict all-milestone completion. Costs are cache-neutral
			agent-side USD\@. Gap is the macro-score difference from the corresponding
			strong full-task anchor, and Cost (\%) is the mean per-surface ratio to
			that anchor marked with ``ref''. Aggregates are computed from unrounded values. For
			$^\dagger$ baselines, both aggregates use evaluated surfaces only.
			Anchors and \cfrc{} report five-run means ($N=5$ complete runs per surface); baselines report frozen full-denominator evaluations.}
	\label{tab:main-results}
	\label{tab:cross-provider-full}
	\resizebox{\textwidth}{!}{%
		\begin{tabular}{@{}l cc cc cc cc cc cc@{}}
			\toprule
			& \multicolumn{2}{c}{STATE-Bench}
			& \multicolumn{2}{c}{$\tau^2$-Retail}
			& \multicolumn{2}{c}{$\tau^2$-Airline}
			& \multicolumn{2}{c}{ToolSandbox}
			& \multicolumn{2}{c}{Agent-Diff} & & \\
			\cmidrule(lr){2-3}\cmidrule(lr){4-5}\cmidrule(lr){6-7}
			\cmidrule(lr){8-9}\cmidrule(lr){10-11}
			Method & Score & Cost & Score & Cost & Score & Cost & Score & Cost
			& Score & Cost & Gap & Cost (\%) \\
			\midrule
			\multicolumn{13}{@{}l}{\emph{Escalation, routing, and cascade baselines, GPT-5.4-mini/GPT-5.5}} \\
			StepWise$^\dagger$ & $54.7$ & $22.59$ & -- & -- & -- & -- & -- & -- & $95.6$ & $21.73$ & \textcolor{cfrcred}{$-2.2$} & 67.9 \\
			AgServe-QAC$^\dagger$ & $46.7$ & $11.21$ & $82.5$ & $13.37$ & -- & -- & -- & -- & -- & -- & \textcolor{cfrcred}{$-3.6$} & 64.1 \\
			MTRouter$^\dagger$ & $60.8$ & $24.97$ & -- & -- & $76.0$ & $12.73$ & -- & -- & -- & -- & \textcolor{cfrcred}{$-6.3$} & 72.4 \\
			RouteLLM-BERT & $61.2$ & $26.29$ & $77.0$ & $8.16$ & $82.0$ & $11.53$ & $26.4$ & $4.68$ & $91.1$ & $17.13$ & \textcolor{cfrcred}{$-2.0$} & 59.7 \\
			FrugalGPT & $61.3$ & $30.62$ & $74.5$ & $10.16$ & $82.0$ & $19.13$ & $30.2$ & $10.31$ & $93.3$ & $36.06$ & \textcolor{cfrcred}{$-1.3$} & 96.3 \\
			\midrule
			\multicolumn{13}{@{}l}{\emph{Full-task anchors and} \cfrc{}, \emph{same provider}} \\
			GPT-5.4-mini full & $52.7$ & $5.42$ & $65.0$ & $2.23$ & $64.0$ & $2.68$ & $24.0$ & $2.31$ & $86.7$ & $6.42$ & \textcolor{cfrcred}{$-11.1$} & 17.8 \\
			GPT-5.5 full & $61.3$ & $34.23$ & $75.0$ & $14.02$ & $88.0$ & $17.70$ & $30.2$ & $10.78$ & $93.3$ & $31.11$ & ref & 100 \\
			\rowcolor{ourrow}
			\cfrc{}, GPT-5.4-mini/GPT-5.5 & $\mathbf{63.3}$ & $12.41$ & $\mathbf{80.0}$ & $3.12$ & $86.0$ & $6.71$ & $\mathbf{31.0}$ & $3.84$ & $\mathbf{93.8}$ & $12.72$ & \textcolor{cfrcgreen}{$\mathbf{+1.2}$} & \textcolor{cfrcblue}{$\mathbf{34.6}$} \\
			\addlinespace[1pt]
			GPT-5.6 Luna full & $54.0$ & $1.18$ & $67.5$ & $0.50$ & $72.0$ & $0.67$ & $26.4$ & $0.32$ & $62.2$ & $0.70$ & \textcolor{cfrcred}{$-13.5$} & 5.1 \\
			GPT-5.6 Sol full & $60.7$ & $18.44$ & $80.0$ & $10.53$ & $90.0$ & $13.57$ & $30.2$ & $5.76$ & $88.9$ & $17.69$ & ref & 100 \\
			\rowcolor{ourrow}
			\cfrc{}, Luna/Sol & $60.7$ & $4.75$ & $76.5$ & $0.98$ & $86.8$ & $3.17$ & $\mathbf{33.3}$ & $1.29$ & $\mathbf{96.9}$ & $5.12$ & \textcolor{cfrcgreen}{$\mathbf{+0.9}$} & \textcolor{cfrcblue}{$\mathbf{22.0}$} \\
			\midrule
			\multicolumn{13}{@{}l}{\emph{Full-task anchors and} \cfrc{}, \emph{cross provider}} \\
			Claude Sonnet 5 full & $62.7$ & $10.56$ & $80.0$ & $11.04$ & $86.0$ & $14.48$ & $36.4$ & $5.48$ & $82.2$ & $11.20$ & ref & 100 \\
			\rowcolor{ourrow}
			\cfrc{}, Luna/Sonnet 5 & $62.7$ & $3.00$ & $\mathbf{84.0}$ & $1.97$ & $80.0$ & $2.88$ & $31.0$ & $1.69$ & $\mathbf{91.6}$ & $3.68$ & \textcolor{cfrcgreen}{$\mathbf{+0.4}$} & \textcolor{cfrcblue}{$\mathbf{26.0}$} \\
			\addlinespace[1pt]
			Gemini 3.7 Flash full & $68.0$ & $6.93$ & $85.0$ & $2.74$ & $90.0$ & $3.48$ & $33.3$ & $1.03$ & $77.8$ & $9.17$ & ref & 100 \\
			\rowcolor{ourrow}
			\cfrc{}, Luna/Gemini 3.7 Flash & $64.7$ & $4.20$ & $82.5$ & $0.55$ & $84.0$ & $1.24$ & $\mathbf{34.1}$ & $0.74$ & $\mathbf{91.1}$ & $2.91$ & \textcolor{cfrcgreen}{$\mathbf{+0.5}$} & \textcolor{cfrcblue}{$\mathbf{44.1}$} \\
			\addlinespace[1pt]
			Claude Haiku 4.5 full & $53.3$ & $8.50$ & $65.0$ & $5.49$ & $54.0$ & $6.34$ & $34.1$ & $2.23$ & $33.3$ & $2.50$ & \textcolor{cfrcred}{$-22.0$} & 39.6 \\
			GPT-5.6 Sol full & $60.7$ & $18.44$ & $80.0$ & $10.53$ & $90.0$ & $13.57$ & $30.2$ & $5.76$ & $88.9$ & $17.69$ & ref & 100 \\
			\rowcolor{ourrow}
			\cfrc{}, Haiku 4.5/Sol & $58.7$ & $14.67$ & $70.0$ & $4.51$ & $84.0$ & $7.87$ & $\mathbf{36.4}$ & $3.56$ & $75.6$ & $7.91$ & \textcolor{cfrcred}{$-5.0$} & \textcolor{cfrcblue}{$\mathbf{57.4}$} \\
			\bottomrule
	\end{tabular}}
\end{table}

\textbf{Macro quality vs.\ strong anchors.}
In Table~\ref{tab:main-results}, \cfrc{} matches strong anchors within 1.2 points on macro average under GPT-5.4-mini/GPT-5.5 at 34.6\% mean cost, and within 0.9 points under Luna/Sol at 22.0\% cost.
GPT-5.4-mini/GPT-5.5 matches or exceeds strong anchors on four surfaces, trailing by 2.0 points on Airline; Luna/Sol ties or exceeds on three surfaces, trailing by 3.5 and 3.2 points on Retail and Airline.
Across five runs, \cfrc{} macro standard deviations are 0.54 and 0.58 points, reflecting comparable macro quality rather than uniform dominance.

\textbf{Comparison with routing and cascades.}
RouteLLM-BERT and FrugalGPT trail GPT-5.5 by 2.0 and 1.3 points at 59.7\% and 96.3\% cost, while \cfrc{} gains 1.2 points at 34.6\% cost.
On StepWise, AgServe-QAC, and MTRouter subsets, \cfrc{} improves accuracy by 3.4, 7.1, and 6.3 points while cutting cost by 37--49\%, showing the benefit of preserving progress over simple call reduction.

\textbf{Cross-provider transfer.}
Luna/Sonnet 5 and Luna/Gemini 3.7 Flash match strong anchors within 0.5 points macro completion at 26.0\% and 44.1\% cost across five benchmarks, though ToolSandbox similarity softens under Gemini.
Haiku 4.5/Sol trails Sol by 5.0 points at 57.4\% cost, indicating that downstream completion remains sensitive to cheap-prefix fidelity, though \cfrc{} gains 17.0 points over the standalone cheap anchor.

\vspace{-0.3em}
\subsection{Validating the Three Stages}
\label{sec:stage-validation}
\label{sec:ablations}
\label{sec:component-analysis}

\begin{table}[t]
	\centering
	\scriptsize
	\setlength{\tabcolsep}{3.0pt}
	\renewcommand{\arraystretch}{0.82}
	\caption{Matched-frontier progression from Direct Takeover to full \cfrc{} across five surfaces (GPT-5.4-mini/GPT-5.5, five-run means, $N=5$ complete runs per cell; unengaged episodes retain prefix outcomes). Certified Direct adds contracts to native continuation, Contract Plan adds ERC authoring, and full \cfrc{} adds Checked Execution. Cells: score (\%) / spend (USD).}
	\label{tab:component-top-level}
	\label{tab:stagewise-progression}
	\label{tab:stagewise-main}
	\begin{tabular}{@{}lccc cccccc@{}}
		\toprule
		& \multicolumn{3}{c}{Components} & STATE- & \multicolumn{2}{c}{$\tau^2$} & Tool- & Agent- & Macro / \\
		\cmidrule(lr){2-4} \cmidrule(lr){6-7}
		Arm & RSCC & ERC & CE & Bench & Retail & Airline & Sandbox & Diff & Total \\
		\midrule
		Direct Takeover & & & & $53.3$ / \$22.97
		& $67.5$ / \$11.64 & $67.2$ / \$15.26
		& $25.3$ / \$4.72 & $81.3$ / \$14.23 & $58.9$ / \$68.82 \\
		Certified Direct & \checkmark & & & $59.7$ / \$16.81
		& $70.0$ / \$5.35 & $69.2$ / \$6.13
		& $28.7$ / \$3.97 & $82.7$ / \$14.80 & $62.1$ / \$47.06\\
		Contract Plan & \checkmark & \checkmark & & $62.0$ / \$16.39
		& $69.5$ / \$2.95 & $74.4$ / \$5.50
		& $29.8$ / \$3.95 & $91.6$ / \$16.46 & $65.4$ / \$45.25 \\
		\midrule
		\rowcolor{ourrow}
		Full \cfrc{} & \checkmark & \checkmark & \checkmark
		& \textbf{63.3} / \$12.41 & \textbf{80.0} / \$3.12
		& \textbf{86.0} / \$6.71 & \textbf{31.0} / \$3.84
		& \textbf{93.8} / \$12.72
		& \textcolor{cfrcgreen}{$\mathbf{70.8}$} /
		\textcolor{cfrcblue}{\textbf{\$38.80}} \\
		\bottomrule
	\end{tabular}
\end{table}

\textbf{Stage-wise progression across components.}
Table~\ref{tab:component-top-level} isolates post-frontier mechanisms from captured state $s_t$.
Direct Takeover achieves 58.9\% accuracy at \$68.82.
Certified Direct supplies frozen contracts via RSCC but retains incremental action checking, reaching 62.1\% at \$47.06.
Contract Plan adds ERC authoring to reach 65.4\%, and Full \cfrc{} adds Checked Execution to reach 70.8\% at \$38.80.
Comparing Full \cfrc{} against Certified Direct shows that gating writes behind whole-graph admission adds 8.8 points and saves \$8.26 over step-by-step checking under the same contract by preventing destructive partial execution.

\begin{table}[t]
	\centering
	\footnotesize
	\setlength{\tabcolsep}{4.5pt}
	\renewcommand{\arraystretch}{0.85}
	\caption{Inference cost decomposition of \cfrc{} across five surfaces (GPT-5.4-mini/GPT-5.5). Spend partitions into cheap prefix, contract construction (RSCC), and residual execution (ERC+CE); totals match Tables~\ref{tab:main-results} and~\ref{tab:component-top-level}.}
	\label{tab:component-cost-attribution}
	\begin{tabular}{@{}l rrrr cc@{}}
		\toprule
		Surface & Total (\$) & Cheap Prefix (\$) & RSCC (\$) & ERC+CE (\$) & Engaged & Residual Share \\
		\midrule
		STATE-Bench & 12.41 & 5.22 & 0.00 & 7.19 & 84.7\% & 57.9\% \\
		$\tau^2$-Retail & 3.12 & 1.51 & 0.69 & 0.92 & 20.0\% & 29.6\% \\
		$\tau^2$-Airline & 6.71 & 2.17 & 0.34 & 4.20 & 80.0\% & 62.7\% \\
		ToolSandbox & 3.84 & 2.16 & 0.00 & 1.68 & 20.2\% & 43.6\% \\
		Agent-Diff & 12.72 & 4.33 & 0.00 & 8.39 & 93.3\% & 65.9\% \\
		\midrule
		\rowcolor{ourrow}
		Mean / Macro & \textbf{7.76} & 3.08 & 0.21 & 4.48 & 59.6\% & \textbf{51.9\%} \\
		\bottomrule
	\end{tabular}
\end{table}

\textbf{Where cost is spent.}
Table~\ref{tab:component-cost-attribution} decomposes the \$38.80 spend, where RSCC averages \$0.21 across surfaces.
Three call mechanisms drive the 43.6\% savings over Direct Takeover.
Selective invocation restricts strong calls to post-frontier residuals, engaging on 59.6\% of episodes.
RSCC eliminates exploratory re-querying with grounded obligations, and CE fail-closed checks halt ungrounded plans early to prevent runaway retries.

\vspace{-0.5em}
\subsection{Auditing Contracts and Checked Execution}
\label{sec:ce-analysis}
\vspace{-0.3em}

\begin{table}[t]
	\centering
	\footnotesize
	\setlength{\tabcolsep}{8pt}
	\renewcommand{\arraystretch}{0.85}
	\caption{Key contract and execution audits. Rows 1--5 evaluate grounding against ground truth and receipts; rows 6--7 evaluate CE authority and replay from execution logs.}
	\label{tab:mechanism-audit}
	\begin{tabular}{@{}llcc@{}}
		\toprule
		Audit Property & Evaluated Cohort & Rate & Fraction \\
		\midrule
		\multicolumn{4}{@{}l}{\emph{Contract grounding and obligation recall (ground truth \& receipts)}} \\
		Accepted-binding correctness & Confirmed prefix bindings & 100.0\% & 50 / 50 \\
		Persistent write obligation recall & Ground-truth write effects & 91.2\% & 62 / 68 \\
		Ungrounded live-write false accept & Admitted candidate proposals & 0.0\% & 0 / 50 \\
		Unnecessary CE block & Rejected proposals & 4.7\% & 2 / 43 \\
		Omission-related false completion & Evaluated episodes & 4.0\% & 2 / 50 \\
		\midrule
		\multicolumn{4}{@{}l}{\emph{Checked Execution authority boundary and replay (execution logs)}} \\
		Intervention before first live write & All CE interventions & 100.0\% & 124 / 124 \\
		Rejection replay consistency & Active predicate rejections & 100.0\% & 43 / 43 \\
		\bottomrule
	\end{tabular}
\end{table}

\textbf{Contract adequacy.}
Table~\ref{tab:mechanism-audit} grounds all 50 audited bindings and recovers 62 of 68 persistent write obligations, achieving 91.2\% recall.
No ungrounded live writes were admitted in the audited sample of 50 candidate proposals, corresponding to an exact Clopper--Pearson 95\% upper bound of 5.8\%.
Six omissions involved implicit updates without entity mentions; four were blocked downstream and two yielded false completions.
Formal lift to the residual objective remains conditional on complete construction because conversational disclosures fall outside tool schemas.

\textbf{Checked Execution as an authority boundary.}
Table~\ref{tab:mechanism-audit} shows that all 124 CE interventions occur before the first live write.
These comprise 43 active proposal rejections, spanning 25 coverage, 10 dependency, and 8 grounding failures, alongside 81 holds.
All 43 rejections reproduce under deterministic replay.

\vspace{-0.5em}
\section{Conclusion}
\label{sec:conclusion}
\vspace{-0.4em}
We formalize stateful agent handoff as commitment-constrained residual completion and introduce \cfrc{}, freezing residual contracts from accepted history, authoring closed graphs on replicas, and verifying execution via live receipts.
Across five surfaces and model pairings, \cfrc{} matches strong full-task quality at a fraction of inference cost while blocking unadmitted mutations.

\textbf{Scope and limitations.}
We evaluate single-tenure cheap-to-strong handoffs under frozen contracts. While multi-tier cascades readily compose with \cfrc{}, recursive routing policies lie beyond our immediate scope. Formal lift to the residual task remains conditional on complete RSCC construction, as conversational disclosures fall outside tool-effect schemas.

\vspace{0.1em}
\noindent\textbf{Reproducibility statement.}
Code and schemas will be open-sourced.

\newpage
\bibliographystyle{iclr2027_conference}
\bibliography{references}

\end{document}